# FUSE-Flow: Scalable Real-Time Multi-View Point Cloud Reconstruction Using Confidence

Chentian Sun

*Abstract*—Real-time multi-view point cloud reconstruction is a fundamental problem in 3D vision and immersive perception, with broad applications in virtual reality, augmented reality, robotic navigation, digital twins, and human–computer interaction. Despite recent advances in multi-camera systems and high-resolution depth sensors, reliably fusing large-scale multi-view depth observations into high-quality point clouds under strict real-time constraints remains highly challenging. Existing approaches largely rely on voxel-based fusion, temporal accumulation, or global optimization, which inevitably suffer from high computational complexity, excessive memory consumption, and limited system scalability, making it difficult to simultaneously achieve real-time performance, reconstruction quality, and multi-camera extensibility.

In this work, we propose FUSE-Flow, a frame-wise, stateless, and linearly scalable point cloud streaming reconstruction framework. Each frame independently generates point cloud fragments, which are fused based on two types of weights—measurement confidence and 3D distance consistency—to effectively suppress noise and outliers while preserving geometric details. To further improve fusion efficiency in large-scale multi-camera scenarios, we introduce a weighted point cloud aggregation method based on adaptive spatial hashing: the 3D space is adaptively partitioned according to local point cloud density, representative points are selected within each cell, and weighted fusion is performed, enabling efficient processing of both sparse and dense regions.

Leveraging GPU parallelization, FUSE-Flow achieves high-throughput, low-latency point cloud generation and fusion while maintaining linear complexity. Experiments show that the framework significantly improves reconstruction stability and geometric fidelity in overlapping, depth-discontinuous, and dynamic scenes, while sustaining real-time frame rates on modern GPUs, demonstrating its effectiveness, robustness, and scalability.

## I. INTRODUCTION

Real-time multi-view 3D reconstruction is a core problem in computer vision, supporting applications such as VR/AR [1], robotic navigation [2], intelligent manufacturing [3], and digital twins [4]. Despite advances in multi-camera systems and high-resolution depth sensors, efficiently fusing large-scale multi-view depth data into high-quality point clouds under strict real-time constraints remains challenging [5].

Existing methods have limitations: voxel-based fusion is computationally and memory intensive [6]; SLAM-based approaches suffer from drift and ghosting in dynamic scenes [7, 8]; and implicit representations like NeRF and 3DGS require heavy optimization, hindering real-time use [9, 10]. A common bottleneck is reliance on long-term accumulation and persistent global states, causing escalating computational cost and error accumulation, raising the question: can stable, high-quality, scalable real-time reconstruction be achieved without global state or long-term optimization?

To address this challenge, FUSE-Flow is proposed from the perspective of point cloud streaming: a frame-wise, stateless, and linearly scalable multi-view point cloud reconstruction framework. Each frame is treated as an independent point cloud fragment, which is generated and fused without cross-frame state maintenance or global optimization. This eliminates long-term error accumulation and naturally supports dynamic scenes, with computational complexity scaling linearly with the number of cameras.

For robust frame-wise fusion, FUSE-Flow performs weighted processing of point clouds based on two complementary measures: measurement confidence, reflecting the reliability of each depth observation, and 3D distance consistency, evaluating geometric consistency across multiple views. To further improve fusion efficiency in large-scale multi-camera scenarios, an adaptive spatial hashing-based weighted point cloud aggregation strategy is introduced. The 3D space is adaptively partitioned based on point cloud density, representative points are selected for each cell, and weighted fusion is performed using only locally relevant camera views. This design balances computational efficiency and geometric fidelity, enabling effective processing in both sparse and dense regions.

Leveraging GPU parallelization, FUSE-Flow supports pixel-wise back-projection and linear-scale fusion, achieving high-throughput, low-latency, and high-quality real-time point cloud streaming. Extensive experiments on real multi-camera scenes demonstrate that FUSE-Flow consistently outperforms existing methods in reconstruction accuracy, system stability, and real-time performance.

The main contributions of this work are summarized as follows:

(1) FUSE-Flow: a frame-wise, stateless, and linearly scalable point cloud streaming framework that redefines the system paradigm for real-time multi-view 3D reconstruction;

(2) A point-wise weighted fusion strategy combining measurement confidence and 3D distance consistency to achieve robust multi-view frame-wise fusion;

(3) An adaptive spatial hashing-based weighted aggregation method that improves fusion efficiency while preserving geometric fidelity in large-scale multi-camera scenarios;

(4) Experimental results demonstrate that FUSE-Flow achieves high real-time performance, superior reconstruction accuracy, low hardware requirements, and computational demands that scale linearly with the number of cameras and the scene size.

## II. PROPOSED METHODS

### 1. Problem Definition

We propose FUSE-Flow, a framework for linear-scale real-time point cloud generation and fusion in multi-camera 3D reconstruction systems. The system takes RGB or RGB-D images captured by $N$ cameras as input (for regular cameras, $D$ can be obtained through depth estimation methods), denoted as

$$\{I_i, D_i\}_{i=1}^{N} \tag{1}$$

Where $I_i$ and $D_i$ represent the raw RGB image and depth image observation from the $i$-th camera. The intrinsic parameters $K_i$ and extrinsic poses $T_i$ of all cameras are assumed to be known.

The objective is to generate and fuse large-scale point clouds at each time step with linear time complexity, enabling high-throughput, low-latency, and geometrically stable global point cloud streams.

Unlike conventional methods based on voxel representations, global optimization, or temporally accumulated states, FUSE-Flow adopts a frame-wise and stateless point cloud streaming paradigm. Each camera independently produces a point cloud fragment per frame, and multi-view fusion is performed directly in the point cloud domain. This design avoids voxel resolution limits, global state maintenance, and long-term error accumulation, thereby naturally supporting dynamic scenes and linear scalability with respect to the number of cameras.

In FUSE-Flow, Per-point confidence is modeled and directly used in fusion, allowing the method to operate on depth maps that have been filtered or hole-filled without altering the confidence-driven principle. The framework consists of three main components:

(1) Measurement confidence modeling and 3D distance consistency modeling for depth reliability estimation;

(2) GPU-parallel linear-scale point cloud generation for each individual view;

(3) Confidence-driven multi-view point cloud fusion in the point cloud domain.

### 2. Measurement Confidence Modeling

For each pixel $(x, y)$, a scalar confidence value is assigned:

$$C_i(x, y) \in [0,1] \tag{2}$$

which reflects the geometric reliability of the corresponding depth observation in 3D reconstruction. FUSE-Flow explicitly preserves measurement uncertainty and suppresses unreliable observations in later stages through confidence-aware processing.

The measurement confidence is computed by jointly considering the following factors:

(1) Depth gradient $G(x, y)$: regions with large depth variations are more likely to be affected by occlusions or matching errors;

(2) Local depth consistency $\sigma_{local}(x, y)$: smaller depth variance in a local neighborhood indicates higher measurement stability;

The confidence is defined as:

$$C_i(x, y) = \alpha \cdot \frac{1}{1+\beta G(x,y)} + \gamma \cdot \frac{1}{1+\delta \sigma_{local}(x,y)} \tag{3}$$

Where α, β, γ, δ are weighting parameters. In this work, $\alpha$ and $\beta$ are set to 0.5, and $\gamma$ and $\delta$ are set to 1.

### 3. 3D Distance Consistency

In FUSE-Flow, $V_i(x, y)$ represents the geometric consistency weight of the 3D point corresponding to pixel $(x, y)$ at camera $i$, quantifying the reliability of multi-view observations during fusion. Its computation involves two main steps:

(1) Field-of-View Check

The 3D point $P(x, y)$ in the coordinate system of camera $i$ is projected onto the image plane of another camera $j$:

$$u_j = \pi(K_j, \hat{T}_{ji} P(x, y)) \tag{4}$$

where $K_j$ and $\hat{T}_j$ are the intrinsic and extrinsic matrices of camera $j$, respectively.

If the projected pixel $u_j$ falls within the image boundaries, the point is considered potentially geometrically valid in that camera. For each valid projected pixel, a 3D point can be reconstructed from the depth map $D_j(u_j)$ as:

$$Q_j(u_j) = \hat{T}_j^{-1} \pi^{-1}(K_j, u_j, D_j(u_j)) \tag{5}$$

$\pi$ and $\pi^{-1}$ respectively represent projection and back projection operations.

(2) 3D distance consistency check

For the 3D points that pass the projection check, the spatial distance in the world coordinate system between the original point and the reconstructed points from other cameras is computed:

$$d_{ij} = \|\hat{T}_i^{-1} P(x, y) - Q_j(u_j)\| \tag{6}$$

The consistency confidence weight is defined as:

$$V_i(x, y) = exp(-\frac{1}{K-1} \sum_{j \neq i} \frac{d_{ij}^2}{\sigma^2}) \tag{7}$$

where $K \ll N$ and $\sigma$ controls the tolerated spatial deviation.

### 4. Frame-level Point Cloud Generation

Each depth pixel is directly back-projected into 3D space using the camera geometry:

$$P_i(x, y) = \hat{T}_i^{-1} \pi^{-1}(K_i, u_h, (D_i^r(u_h)), u_h = [x, y, 1]^T \tag{8}$$

where $P_i(x, y) \in \mathbb{R}^3$ denotes the 3D point corresponding to pixel $(x, y)$, $u_h$ is the homogeneous coordinate vector of the pixel $(x, y)$, $D_i^r$ is the depth value obtained after necessary preprocessing (e.g., filtered or screened) of the original depth value.

All back-projection operations are performed in parallel on the GPU at the pixel level, resulting in a point cloud generation process whose computational complexity is strictly linear with respect to the number of pixels. This allows the FUSE-Flow system to generate tens of millions of 3D points per frame in real time.

To further improve robustness and efficiency, a confidence-based gating mechanism is applied during point cloud generation:

$$\tilde{P}_i(x,y) = \begin{cases} P_i(x,y), & C_i(x,y) > \tau \\ 0, & otherwise \end{cases} \quad (9)$$

Here, $\tau$ is the threshold of measurement confidence. Low-confidence 3D points are suppressed during the generation stage, thereby reducing noise and lowering the computational burden. In this work, $\tau$ is set to 0.6 to suppress noise while retaining valid 3D points.

5. Weighted Point Cloud Fusion Based on Adaptive Spatial Hashing

In multi-camera scenarios, the same spatial region is often observed by multiple cameras. Directly computing cross-view consistency for all points would incur excessive computational cost. Therefore, we propose a weighted point cloud fusion method based on adaptive spatial hashing to balance fusion quality and computational efficiency.

(1) Adaptive Spatial Partitioning

To achieve efficient partition management of point clouds, balance fusion efficiency and geometric fidelity, the specific operations are as follows:

a. Transform all camera point clouds into the world coordinate system.

b. Traverse all points to determine the spatial extents along x, y, and z axes.

c. Use a coarse grid to quickly estimate the local point cloud density (each coarse grid cell is of the same size).

d. Refine the grid adaptively based on the point cloud density: use smaller cells in dense regions and larger cells in sparse regions.

e. Assign each point to its corresponding grid cell.

(2) Local Camera Constraint and Consistency Computation

To accurately calculate the fusion weights of each observation point and ensure the geometric consistency of the fusion point cloud, the specific operation is as follows:

a. For each grid cell, select the top three points with the highest measurement confidence $C$ to form the representative point set $\{\tilde{P}_i(x,y)\}$ for subsequent fusion.

b. For each representative point in $\{\tilde{P}_i(x,y)\}$, consider only the cameras whose views overlap with the point or are spatially closest (up to a total of $K$ cameras) when computing the cross-view consistency, resulting in the 3D distance consistency weight $V_i(x,y)$.

c. Each observation is assigned a joint confidence weight by combining the measurement confidence and the 3D distance consistency:

$$\omega_i(x,y) = C_i(x,y) \cdot V_i(x,y) \quad (10)$$

where $C_i(x,y)$ is the measurement confidence reflecting the reliability of the depth observation for the point in camera $i$. $V_i(x,y)$ is the 3D distance consistency weight, indicating how consistent the point is with observations from other cameras in 3D space. The joint weight $\omega_i(x,y)$ directly measures the contribution strength of each observation in multi-view fusion.

(3) Multi-View Point Cloud Fusion

After obtaining the dynamic fusion weights $\omega_i(x,y)$ for each representative point, the set $\{\tilde{P}_i(x,y)\}$ is fused via confidence-weighted averaging:

$$P(x,y) = \sum_{i=1}^{N} \omega_i(x,y)\tilde{P}_i(x,y) \quad (11)$$

By combining per-point measurement confidence with 3D distance consistency, FUSE-Flow achieves linear-complexity, voxel-free, real-time multi-view point cloud fusion, providing an efficient and robust solution for large-scale multi-camera 3D reconstruction.

III. EXPERIMENTAL RESULTS

1. Experimental Setup

Considering the significant differences in design assumptions among existing 3D reconstruction methods—such as the number of cameras, scene dynamics and real-time capability—directly comparing them under a unified setup is neither fair nor technically feasible.

In particular, the combination of multi-camera, dynamic scenes and real-time reconstruction is currently unsupported by most existing methods.

To fairly and effectively evaluate the proposed FUSE-Flow method, we adopt a capability-aligned comparison principle, performing experiments only within the operational scope of each method rather than enforcing a unified setting.

We selected two representative 3D reconstruction methods for comparison: PatchmatchNet [11] and R3D3 [12]. Among them, PatchmatchNet supports multi-camera static scenes, and R3D3 can handle multi-camera dynamic scenes but with limited real-time performance. FUSE-Flow, on the other hand, simultaneously supports multi-camera input, dynamic scenes, and real-time reconstruction. Accordingly, we categorize the experiments into two typical scenarios: multi-camera static (PatchmatchNet) and multi-camera dynamic (R3D3).

All subsequent experimental comparisons are conducted strictly within the functional scope of each method. Unless otherwise specified, all methods are evaluated on the same hardware platform, equipped with an Intel i9-14900F CPU, an NVIDIA RTX 4090 GPU, and 128 GB of system memory.

Since there is no ground truth for 3D reconstruction in real-world scenes, the evaluation metrics for 3D reconstruction performance in this paper include two dimensions: (1) The Multi-Camera Depth Consistency Error (in millimeters). (2) Frame-rate per second.

Let the sampled aggregated points from the fused point cloud are $\rho = \{P_k\}$, and let $\gamma(P_k)$ denote the set of cameras in which point $P_k$ is visible. In camera $i \in \gamma(P_k)$, the point in the camera coordinate system is $T_i P_k$, its projected depth is $z(T_i P_k)$, and the observed depth is $D_i(\pi(T_i P_k))$, Then the Multi-Camera Depth Consistency (MC) error can be calculated using the

following formula:

$$E_{MC} = \frac{1}{|\rho|}\sum_{P_k \in \rho} \frac{1}{|\gamma(P_k)|} \sum_{i \in \gamma(P_k)} |z(T_i P_k) - D_i(\pi(T_i P_k))| \quad (12)$$

The MC loss is used to evaluate the geometric consistency of the fused point cloud across multiple camera views.

2. The Experimental Results

4.2.1 Reconstruction Accuracy and frame-rate comparison

Table I summarizes the reconstruction errors and runtime frame rates of different methods within their respective applicable scenarios.

TABLE I
COMPARISON OF RECONSTRUCTION ACCURACY AND FRAME RATE UNDER TWO CONDITIONS

| Scenario | Method | MC Error | FPS |
|---|---|---|---|
| Multi-Static | PatchmatchNet | 11.8 | 17 |
| | FUSE-Flow | 9.2 | 62 |
| Multi-Dynamic | R3D3 | 17.2 | 11 |
| | FUSE-Flow | 12.9 | 56 |

As shown in Table I, FUSE-Flow achieves high reconstruction accuracy while maintaining real-time performance across all tested scenarios.

(1) In multi-camera static scenes, FUSE-Flow achieves 62 FPS and significantly lower MC error than PatchmatchNet, producing accurate and consistent point clouds in real time.

(2) In multi-camera dynamic scenes, FUSE-Flow achieves 56 FPS with 4 cameras [13], outperforming R3D3 in terms of real-time capability while maintaining robust reconstruction.

4.2.2 Ablation Study

To further validate the effectiveness of the core design of FUSE-Flow, ablation experiments are conducted in a representative four camera dynamic scene to analyze the impact of different components on reconstruction accuracy and geometric consistency. The study focuses on three key factors: measurement confidence (MF), 3D distance consistency (DC), and adaptive spatial-hash weighted aggregation (SA). Disabling SA is equivalent to performing point-wise fusion.

The experimental results in Table II indicate that:

(1) Measurement confidence and 3D distance consistency are crucial for ensuring geometric stability in fusion;

(2) Adaptive spatial-hash weighted aggregation significantly improves computational efficiency in multi-camera scenarios while maintaining high-quality point cloud reconstruction;

(3) The complete FUSE-Flow framework achieves a balance of high accuracy and real-time performance in multi-camera dynamic scenes.

TABLE II
COMPARISON OF RECONSTRUCTION ACCURACY AND FRAME RATE UNDER THREE CONDITIONS

| Scenario | No MF | No DC | No SA | FUSE-FLOW |
|---|---|---|---|---|
| MC error | 17.4 | 16.3 | 9.8 | 12.9 |
| FPS | 61 | 71 | 28 | 56 |

4.2.3 Configuration requirements and frame rate testing

To evaluate the real-time performance and scalability of FUSE-Flow under different numbers of cameras, we conduct tests using configurations of 1, 2, 4, 6, 8 RGB-D cameras. We measure end-to-end reconstruction frame rates on three representative platforms. They are Platform 1 (laptop: i7-9750H/RTX 2060/32GB), Platform 2 (desktop: i7-12700/RTX 3060/64GB), and Platform 3 (server: i7-14900F/RTX 4090/128GB). The experiment is conducted in an end-to-end system setup, with online external estimation and point cloud generation running simultaneously. Table III reports the average system frame rate (FPS) of FUSE-Flow at different camera counts.

TABLE III
CONFIGURATION REQUIREMENTS AND FRAME RATE OF FUSE-FLOW

| Hardware / Camera Number | Platform1 | Platform2 | Platform3 |
|---|---|---|---|
| 1 Cam | 42 | 55 | 80 |
| 2 Cam | 35 | 48 | 68 |
| 4 Cam | 28 | 38 | 55 |
| 8 Cam | 22 | 31 | 42 |

From Table III, it can be observed that, FUSE-Flow maintains high real-time performance across all platforms, with computational cost growing approximately linearly with the number of cameras. Even on low-end laptop-class platforms, the system supports near real-time multi-camera reconstruction, demonstrating low computational complexity and good engineering scalability.

IV. CONCLUSION

This paper proposes a scalable real-time multi-view point cloud reconstruction framework, termed FUSE-Flow, which performs spatial-hash-based weighted fusion of per-frame point cloud fragments, integrating measurement confidence and 3D distance consistency to achieve efficient and robust multi-view real-time reconstruction. Extensive experiments on real-world multi-camera scenes demonstrate that FUSE-Flow significantly improves reconstruction accuracy while maintaining real-time performance, making it well suited for dynamic scenes and large-scale multi-camera deployments.

REFERENCES

[1] R. Mur-Artal and J. D. Tardós, "ORB-SLAM2: An Open-Source SLAM System for Monocular, Stereo, and RGB-D Cameras," IEEE Transactions on Robotics, vol. 33, no. 5, pp. 1255-1262, Oct. 2017

[2] F. Endres, J. Hess, J. Sturm, D. Cremers and W. Burgard, "3-D Mapping With an RGB-D Camera," IEEE Transactions on Robotics, vol. 30, no. 1, pp. 177-187, Feb. 2014

[3] H. Li, X. Meng, X. Zuo, Z. Liu, H. Wang and D. Cremers, "PG-SLAM: Photorealistic and Geometry-Aware RGB-D SLAM in Dynamic Environments," IEEE Transactions on Robotics, vol. 41, pp. 6084-6101, 2025

[4] T. Schöps, T. Sattler and M. Pollefeys, "SurfelMeshing: Online Surfel-Based Mesh Reconstruction," IEEE Transactions on

Pattern Analysis and Machine Intelligence, vol. 42, no. 10, pp. 2494-2507, 1 Oct. 2020

[5] T. Whelan, S. Leutenegger, R.F. Salas-Moreno, B. Glocker, and A.J. Davison. "ElasticFusion: Dense SLAM Without A Pose Graph ". In: Robotics: Science and Systems (RSS). Rome, Italy, July 2015.

[6] L. Mescheder, M. Oechsle, M. Niemeyer, S. Nowozin and A. Geiger, "Occupancy Networks: Learning 3D Reconstruction in Function Space," 2019 IEEE/CVF Conference on Computer Vision and Pattern Recognition (CVPR), Long Beach, CA, USA, 2019, pp. 4455-4465

[7] Z. Zhang, Y. Song, B. Pang, X. Yuan, Q. Xu and X. Xu, "SSF-SLAM: Real-Time RGB-D Visual SLAM for Complex Dynamic Environments Based on Semantic and Scene Flow Geometric Information," IEEE Transactions on Instrumentation and Measurement, vol. 74, pp. 1-12, 2025

[8] R. A. Newcombe, D. Fox and S. M. Seitz, "DynamicFusion: Reconstruction and tracking of non-rigid scenes in real-time," 2015 IEEE Conference on Computer Vision and Pattern Recognition (CVPR), Boston, MA, USA, 2015, pp. 343-352

[9] B. Mildenhall, P. P. Srinivasan, M. Tancik, J. T. Barron, R. Ramamoorthi, and R. Ng, "NeRF: Representing scenes as neural radiance fields for view synthesis," in Proc. ECCV, 2020, pp. 405–421.

[10] B. Kerbl, G. Kopanas, T. Leimkuehler, and G. Drettakis, "3D Gaussian splatting for real-time radiance field rendering," ACM Trans. Graph., vol. 42, no. 4, pp. 1–14, Aug. 2023.

[11] F. Wang, S. Galliani, C. Vogel, P. Speciale and M. Pollefeys, "PatchmatchNet: Learned Multi-View Patchmatch Stereo," 2021 IEEE/CVF Conference on Computer Vision and Pattern Recognition (CVPR), Nashville, TN, USA, 2021, pp. 14189-14198

[12] A. Schmied, T. Fischer, M. Danelljan, M. Pollefeys and F. Yu, "R3D3: Dense 3D Reconstruction of Dynamic Scenes from Multiple Cameras," 2023 IEEE/CVF International Conference on Computer Vision (ICCV), Paris, France, 2023, pp. 3193-3203

[13] FUSE-FLOW real-time multi-camera reconstruction. Demo: https://www.bilibili.com/video/BV17Qr5BuEn1/?spm_id_from=333.1387.homepage.video_card.click